\documentclass{article}

    \PassOptionsToPackage{numbers, compress}{natbib}

\usepackage[final]{neurips_2022}

\usepackage{amsmath,amssymb}
\usepackage[utf8]{inputenc} 
\usepackage[T1]{fontenc}    
\usepackage{hyperref}       
\usepackage{url}            
\usepackage{booktabs}       
\usepackage{amsfonts}       
\usepackage{nicefrac}       
\usepackage{microtype}      
\usepackage{xcolor}         

\usepackage{graphicx}
\usepackage{wrapfig,lipsum}
\usepackage{bbm}
\usepackage{multicol}

\usepackage{booktabs}
\usepackage{amsfonts}
\usepackage{siunitx}
\usepackage{multirow}

\usepackage{amssymb}
\usepackage{pifont}
\newcommand{\cmark}{\text{\ding{51}}}%
\newcommand{\xmark}{\text{\ding{55}}}%

\title{SL3D: Self-supervised-Self-labeled 3D Recognition}

\author{%
  Fernando Julio Cendra\\
  The University of Hong Kong\\
  \& \\
  TCL AI Lab\\
  \texttt{fcendra@connect.hku.hk}\\
  \And
  Lan Ma\\
  TCL AI Lab\\
  \texttt{rubyma@tcl.com}\\
  \AND
  Jiajun Shen\\
  TCL AI Lab\\
  \texttt{shenjiajun90@gmail.com}\\
  \And
  Xiaojuan Qi\\
  The University of Hong Kong\\
  \texttt{xjqi@eee.hku.hk}\\
}

\begin{document}

\maketitle

\begin{abstract}
  Deep learning has attained remarkable success in many 3D visual recognition tasks, including shape classification, object detection, and semantic segmentation.
  However, many of these results rely on manually collecting densely annotated real-world 3D data, which is highly time-consuming and expensive to obtain, limiting the scalability of 3D recognition tasks. Thus, we study unsupervised 3D recognition and propose a \textbf{S}elf-supervised-Self-\textbf{L}abeled \textbf{3D} Recognition (SL3D) framework. SL3D simultaneously solves two coupled objectives, \emph{i.e.,} clustering  and learning feature representation to generate pseudo-labeled data for unsupervised 3D recognition. SL3D is a generic framework and can be applied to solve different 3D recognition tasks, including classification, object detection, and semantic segmentation. Extensive experiments demonstrate its effectiveness. Code is available at \url{https://github.com/fcendra/sl3d}.
\end{abstract}

\section{Introduction}

3D object recognition is a fundamental problem in computer vision that aims to process given visual data to generate high-level understandings with many applications in robotics, autonomous driving, and virtual reality. Some representative recognition tasks include object classification, detection, and semantic segmentation. Although 3D data acquisition has been convenient and inexpensive with the advancement of 3D sensing technologies, annotating 3D labels is still laborious, inconvenient, and time-consuming, which limits the scalability of many recognition tasks. For example, on average, it takes $22.3$ minutes to annotate the ground-truth data for a single 3D indoor scene data sample on ScanNet \cite{dai2017scannet} dataset, and nearly 500 annotators were involved in constructing the dataset. This problem directly causes a hindrance towards unleashing the full potential of deep learning methods, as the key strength of deep learning methods is their ability to learn hierarchical features from a large amount of annotated training data. 
The unavailability of large-scale labeled data has been one of the most significant factors preventing the community from scaling 3D recognition tasks. We review the related works in greater detail in Sec. \ref{sec:related}.

To this end, we explore the potential of unsupervised 3D recognition, which does not require annotations for model training and thus bypasses the limitations of data annotations. To our best knowledge, this has rarely been  explored in the 3D domain. We propose the SL3D learning paradigm, which tackles 3D recognition problem by simultaneously self-labeling and learning feature representation from unlabeled object-level point cloud data. Inspired by SeLa \cite{asano2020self}, SL3D generates pseudo labels by simultaneously clustering with an equipartition constraint (i.e., each cluster is assigned with an approximately equal number of samples) and learning feature representations from clustered data. Clustering and representation learning is conducted iteratively to finally deliver high-quality pseudo labeled data. Furthermore, these pseudo labeled data obtained from SL3D can be used as supervision for several downstream tasks such as 3D object classification, 3D object detection, and 3D semantic segmentation to achieve unsupervised 3D recognition. As object detection and semantic segmentation tasks deal with scene-level point cloud data, a selective search algorithm is used to generate bounding box proposals from 3D indoor dataset and transform them into object-level point cloud.

We conduct extensive experiments on ModelNet40 dataset \cite{wu20153d} for the object classification task, and ScanNet \cite{dai2017scannet} dataset for both object detection and semantic segmentation tasks.
The experimental results show that our method achieves promising results for unsupervised 3D object detection and 3D semantic segmentation. Moreover, through extensive experiments, we show that our Unsupervised 3D object detection achieves on-par results compared with the base benchmark model for Weakly-supervised 3D object detection. Our contributions can be summarized as below:
\setlength{\parskip}{0.0em}
\begin{itemize}
    \item{We propose SL3D for unsupervised 3D recognition, a generic framework for 3D object classification, semantic segmentation, and object detection.} 
    \item{We conduct extensive experiments to verify the effectiveness of our framework on three challenging 3D recognition tasks.}
\end{itemize}
\setlength{\parskip}{0.0em}
The rest of the paper is organized as follows. Section \hyperref[sec:method]{2} gives an overview of our method. Section \hyperref[sec:exp]{3} reports the experimental results of our SL3D model on downstream tasks. Finally, in section \hyperref[sec:end]{4}, the paper ends with a conclusion.

\section{Method}
\label{sec:method}
The core task of our framework, summarized in Sec. \ref{sec:illustration}, is to generate pseudo labels from given point cloud data which can be used as training data for downstream 3D object recognition tasks, including object classification, semantic segmentation, and object detection. To achieve our goal, we design a simultaneous self-supervised-self-labeled (SL3D) method that generates labels through clustering from sets of object-level point clouds $\{ P^O_{j} \mid j=1, \dots, |P^O| \}$, {where $P_j^O$ is a vector consists of $(x,y,z)$ coordinates representing object-level point cloud $j$.} Given scene-level point cloud data $\{ P^S_{i} \mid i=1, \dots, m \}$ as the input, a bottom-up 3D geometric selective search object proposal algorithm \cite{ren2021wypr} is employed to generate sets of $P_j^O$ for each $P^S_{i}$, where $i$ corresponds to the scene index. Afterward, the scene point cloud is represented as a set of object-level proposals. In the following, we will first elaborate the SL3D model for generating high-quality pseudo labels. Then, we will explain how the pseudo labeled data are used for solving downstream tasks.

\subsection{Simultaneous self-supervised-Self-labeled model}

Given a set of object-level  point clouds X = $\{ x \mid \forall x \in P^O\}$ which are either data samples from an object-level classification dataset such as ModelNet40 \cite{wu20153d} dataset or proposals generated from a scene-level point cloud dataset such as ScanNet \cite{dai2017scannet} dataset, we develop our self-supervised-self-labeled (SL3D) framework inspired by a 2D-based self-supervised model proposed by Asano \textit{et al.}\cite{asano2020self}. The key idea behind this method is to simultaneously train a feature representation learning model that learns from a joint distribution generated by the clustering model. 
Therefore, the model is divided into two coupled parts \textit{i.e.,} \textbf{1. Clustering model:} Given features from a network, a  clustering model is developed to assign data to a pseudo cluster with an equipartition constraint to ensure the data distribute equally amongst the pseudo classes. \textbf{2. Feature representation learning:} The network is supervised by minimizing the cross-entropy loss between pseudo label distribution Q and the distribution P obtained from the model predictions. The two models work collaboratively to update the backbone network $f(\cdot)$ which can, in turn, generate better pseudo labels.

\paragraph{Clustering model:} Given joint probability distribution $P$ obtained from Eq.(\ref{eq1}) where each element represents the probability of a sample belonging to a class, a naive approach would be to assign a data sample to the class with the largest probability
and using it as supervision for feature representation learning. However, this would lead to a degeneracy solution as all the samples $X$ have the potential to be clustered into one pseudo class. 
\begin{equation}
    P_{y_{i}} = \frac{1}{N}\mbox{softmax}(h \circ f(x_{i})) \label{eq1}
\end{equation} 
\setlength{\parskip}{0.0em}
To avoid this undesirable situation, we adopt the approach proposed by Asano \textit{et al.}\cite{asano2020self}, which employs an equipartition constraint to ensure $X$ to be assigned uniformly to each pseudo class, i.e., balanced pseudo clusters. 
Initially, we set $P$ to be a $K\times N$ joint probabilities matrix obtained following Eq.\eqref{eq1}, where $N$ is the number of object-level data samples in $X$, $K$ is the number of pseudo classes (groups), $h$ is the classifier head with a length of $K$, and $\mbox{softmax}(\cdot)$ is the normalization operation. Similarly, we also initialize label assignment matrix $Q$ to be $K\times N$. Each element $Q_{ij}$ will be used to represent the pseudo probability of a sample $j$ belonging to a group $i$ following the equipartition constraint.

\paragraph{Feature representation learning model:} Given probability distribution $P$ label assignment matrix $Q$ obtained from the clustering model, we train a representation learning model by minimizing the cross-entropy loss $\hbox{CE}(\cdot,\cdot)$ between Q and P. Thus, the objective function of this model can be written as $\min_{}  \langle Q, -\log P \rangle$, where $\langle \cdot \rangle$ is the Frobenius dot product operation between two matrices. Finally, a fast version of Sinkhorn-Knopp algorithm \cite{cuturi2013sinkhorn} is used to optimize this objective function efficiently.

\setlength{\parskip}{0.0em}

Thus, we see that by clustering, and learning representations simultaneously, minimizing the objective function is the same way as maximizing the mutual information between the label and data index $i$ while still maintaining equipartition among the $N$ data. To obtain the pseudo labels $Y$ for further use in 3D recognition tasks, we can simply take the argument that gives the maximum value for every data index in table Q, this can be formulated as $Y = \{ y \mid \forall y \in \hbox{argmax}(Q_{*,i}), \forall i \in (1,\dots, N)\}$.

\subsection{Downstream tasks}

As the downstream task methods are not the core components of SL3D, each task adopts the model from existing 3D object recognition works. As each downstream task requires different types of input training data, another pre-processing task is needed to transform SL3D dataset for each 3D recognition task. The 3D object classification task requires object-level point cloud data and its corresponding class label. The class labels can be obtained from pseudo label generated by SL3D. For classifier model, we use PointNet++ \cite{qi2017pointnetplusplus} as our baseline model. For 3D indoor scene recognition, we evaluate both 3D object detection and 3D semantic segmentation. 3D object detection task requires the semantic labels and bounding box location of each objects, we can directly get the semantic labels from SL3D pseudo labels and obtain the bounding box location from the proposals generated by GSS. After obtaining the object detection dataset from SL3D model, we train VoteNet \cite{qi2019deep} as our object detection model. For 3D semantic segmentation task, as SL3D does not have any point-wise labels, we use the obtained proposal boxes to map and assign all points inside the box with the corresponding SL3D pseudo label. After obtaining the 3D semantic segmentation dataset from SL3D, we use PointNet++ \cite{qi2017pointnetplusplus} semantic segmentation model and train it with our pseudo labeled data.

\section{Experiment}
\label{sec:exp}
We evaluate and compare SL3D framework with the state-of-the-art weakly supervised and supervised methods on three 3D object recognition tasks,i.e., 3D object classification on ModelNet40 \cite{wu20153d} dataset, 3D object detection and semantic segmentation benchmarks on ScanNet \cite{dai2017scannet} dataset. We also present our implementation details in Sec. \ref{sec:implementation}, and ablation experiments in Sec. \ref{sec:ablation}.

\subsection{Dataset} 
For 3D classification tasks, we use the ModelNet40 dataset. It contains $12,311$ object-level point cloud CAD models of $40$ categories. The dataset is divided into a training set ($9,843$ objects) and a testing set ($2,468$ objects). For 3D object detection and 3D semantic segmentation tasks, we use real 3D scene point cloud data, ScanNet \cite{dai2017scannet} dataset. ScanNet consists of $1613$ scene-level indoor scene point cloud data annotated with $20$ semantic classes. The dataset is divided into training ($1201$ scenes), validation ($312$ scenes), and testing ($100$ scenes) sets. 

\setlength{\parskip}{0.0em}

\subsection{Main Results}

Here we present our unsupervised 3D recognition model performance for classification, object detection, and semantic segmentation tasks. We use mean average precision ($\hbox{mAP}@0.25$) for detection task. Moreover, for segmentation task, we use mean intersection over union ($\hbox{mIoU}$) to measure the similarity of the point-level semantic ground truth with our SL3D prediction result. 
As there are no prior works in unsupervised 3D indoor-scene recognition, we show other learning methods, i.e., weakly-supervised and supervised methods, as a reference. The evaluation results are shown in Table \ref{table:2} for classification, Table \ref{table:3} for object detection, and Table \ref{table:4} for semantic segmentation. We define our model settings to be $SL3D^{\{\dagger,\ddagger\}}_{K}$, where $\dagger$ denotes SL3D with PointNet++ backbone, $\ddagger$ denotes SL3D with Point Transformer backbone, and $K$ denotes the number of pseudo classes.
\setlength{\parskip}{0.0em}

\paragraph{Unsupervised 3D object classification on ModelNet40 dataset:}As Table \ref{table:2} shows, our model performs better with the increased number of pseudo classes. Its accuracy gains around $9.7\%$ $(\pm 2.9)$ improvement for every doubling of pseudo class number. Moreover, compared with the baseline and SOTA supervised classification models, our best model (with an accuracy of $77.2 \%$) achieves accuracy differences of $14.7$ and $17.3$, respectively.\textbf{ Unsupervised 3D object detection on ScanNet dataset: } As shown in Table \ref{table:3}, our model obtains decent results in this unsupervised setting. It is reasonable that our model underperforms compared with other models in the weakly supervised or the strongly supervised setting.
Still, our model (with Point Transformer as the backbone architecture and $400$ pseudo classes) achieves competitive results compared to MIL-det \cite{ren2021wypr}, a weakly-supervised object detection-based model. We also evaluate our model under the same configuration on SL3D object detection validation set, i.e., the labels are obtained from SL3D model, and the model achieves an \hbox{$mAP$} of $19.1\%$. 
This performance improvement implies that there is a discrepancy between object groups discovered by SL3D and the ground-truth annotated categories, which leaves room for further improvements. \textbf{ Unsupervised 3D semantic segmentation on ScanNet dataset: } As shown in Table \ref{table:4}, we show that our unsupervised semantic segmentation model performs best with Point Transformer backbone and $400$ pseudo classes and achieves $10.4\%$ of $mIoU$. We observe that increasing the number of pseudo classes improves the overall semantic segmentation results.

\begin{table}[htbp]
    \begin{minipage}{.35\linewidth}
        \setlength{\parskip}{0.0em}
        \centering
        \parbox{6cm}{\caption{3D object classification results on ModelNet40 dataset from different level of supervision.}\label{table:2}}
        \scalebox{0.65}{
        \begin{tabular}{S S S S S} \toprule
            {No.} & {Method} & {No. of class obtained} & {Overall Acc($\%$)} \\ \midrule
            \multicolumn{ 4}{S}{\textbf{Unsupervised methods}} \\ \midrule
            {1}  & {$SL3D^{\ddagger}_{100}$ (ours)}  & {26} & {38.7} \\
            {2}  & {$SL3D^{\ddagger}_{200}$ (ours)}  & {33} & {48.9} \\
            {3}  & {$SL3D^{\ddagger}_{400}$ (ours)}  & {38} & {59.8} \\
            {4}  & {$SL3D^{\ddagger}_{800}$ (ours)}  & {40} & {72.5} \\
            {4}  & {$SL3D^{\ddagger}_{1600}$ (ours)}  & {40} & {\textbf{77.2}} \\ \midrule
            \multicolumn{4}{S}{\textbf{Supervised methods}} \\ \midrule
            {5}  & {PointNet++ \cite{qi2017pointnetplusplus}}  & {40} & {91.9} \\
            {6}  & {PointMLP \cite{ma2022rethinking}}  & {40} & {94.5} \\\bottomrule
            
        \end{tabular}
        }
    \end{minipage}%
    \begin{minipage}{.8\linewidth}
        \centering
        \parbox{7cm}{\caption{3D object detection results ($\hbox{mAP}@0.25$) task on ScanNet validation set from different level of supervision.}\label{table:3}}
        \setlength{\parskip}{0.5em}
        \centering
        \scalebox{0.6}{
        \begin{tabular}{S S S S S} \toprule
            {No.} & {Method} & {No. of class obtained} & {Split} & {$\hbox{mAP}@0.25$} \\ \midrule
            \multicolumn{ 5}{S}{\textbf{Unsupervised methods}} \\ \midrule
            {1}  & {$SL3D^{\dagger}_{50}$ (ours)} & {7} & {Val(SL3D)/Val} & {18.6/4.6} \\
            {2}  & {$SL3D^{\ddagger}_{100}$ (ours)}  & {12} & {Val(SL3D)/Val} & {17.8/7.6} \\
            {3}  & {$SL3D^{\ddagger}_{200}$ (ours)}  & {15} & {Val(SL3D)/Val} & {20.3/7.9} \\
            {4}  & {$SL3D^{\ddagger}_{400}$ (ours)}  & {18} & {Val(SL3D)/Val} & {19.1/\textbf{9.3}} \\\midrule
            \multicolumn{ 5}{S}{\textbf{Weakly-supervised methods}} \\ \midrule
            {5}  & {MIL-det \cite{ren2021wypr}}  & {18} & {Val}  & {9.6} \\
            {6}  & {WyPR \cite{ren2021wypr}}  & {18} & {Val} & {18.3} \\
            {7}  & {VoteNet + WS3D \cite{meng2020weakly}} & {18} & {Val} & {18.4} \\
            {8}  & {WyPR + prior \cite{ren2021wypr}}  & {18} & {Val} & {19.7} \\
            {9}  & {VoteNet + ${BR}_{P}$ \cite{xu2022back}} & {18} & {Val} & {31.2} \\ \midrule
            \multicolumn{ 5}{S}{\textbf{Supervised methods}} \\ \midrule
            {10}  & {F-PointNet \cite{qi2018frustum}}  & {18} & {Val}  & {19.8} \\ 
            {11}  & {GSPN \cite{yi2019gspn}}  & {18} & {Val}  & {30.6} \\ 
            {12}  & {VoteNet \cite{qi2019deep}}  & {18} & {Val}  & {58.6} \\
            {13}  & {RBGNet \cite{wang2022rbgnet}}  & {18} & {Val}  & {70.6} \\ \bottomrule
            \end{tabular}
            }
        \end{minipage}
    \end{table}

\setlength{\parskip}{0.0em}

\begin{table}[!ht]
\caption{3D semantic segmentation results on ScanNet dataset from different level of supervision. Val(SL3D) denotes the SL3D semantic segmentation validation set, and Val(sub-cloud) denotes the PCAM \cite{mcever2020pcams}  and MPRM \cite{wei2020multi} validation sets with sub-cloud supervision.}\setlength{\parskip}{0.0em}
\centering
\scalebox{0.67}{
\begin{tabular}{S S S S S} \toprule
    {No.} & {Method} & {No. of class obtained} & {Split} & {$\hbox{mIoU}$} \\ \midrule
    \multicolumn{ 5}{S}{\textbf{Unsupervised methods}} \\ \midrule
    {1}  & {$SL3D^{\dagger}_{50}$ (ours)} & {9} & {Train/Val(SL3D)/Test} & {60.2/32.9/5.8} \\
    {2}  & {$SL3D^{\ddagger}_{100}$ (ours)}  & {14} & {Train/Val(SL3D)/Test}  & {57.3/26.6/8.4} \\
    {3}  & {$SL3D^{\dagger}_{400}$ (ours)}  & {18} & {Train/Val(SL3D)/Test}  & {56.1/28.5/8.5} \\
    {4}  & {$SL3D^{\ddagger}_{400}$ (ours)}  & {20} & {Train/Val(SL3D)/Test}  & {55.1/25.3/9.2} \\
    {5}  & {$SL3D^{\ddagger}_{800}$ (ours)}  & {20} & {Train/Val(SL3D)/Test}  & {53.6/24.7/10.5} \\ \midrule
    \multicolumn{5}{S}{\textbf{Weakly-supervised methods}} \\ \midrule
    {6}  & {PCAM \cite{mcever2020pcams}} & {20} & {Train/Val(sub-cloud)} & {22.1/28.1} \\
    {7}  & {MPRM \cite{wei2020multi}} & {20} & {Train/Val(sub-cloud)}  & {24.4/41.0} \\
    {8}  & {WyPR \cite{ren2021wypr}} & {20} & {Train/Val/Test}  & {30.7/29.6/24.0} \\
    {9}  & {WyPR + prior \cite{ren2021wypr}} & {20} & {Val}  & {31.1} \\ \midrule
    \multicolumn{5}{S}{\textbf{Supervised methods}} \\ \midrule
    {10}  & {PointNet++ \cite{qi2017pointnetplusplus}} & {20} & {Test}  & {33.9} \\ \bottomrule

\end{tabular}
}
\label{table:4}
\end{table}

\section{Conclusions}
\label{sec:end}

SL3D framework has been proposed in this paper. Given a 3D object-level point cloud data (optionally for scene-level point cloud data, a geometric selective search algorithm is used to create object-level point cloud data), These set of point clouds are then fetched into the simultaneous self-supervised-self-labeled learning model to acquire pseudo labels. 
The pseudo labeled data are further used to supervise different 3D recognition model training including classification, object detection, and semantic segmentation. Our experimental results indicate that this model fulfills our main research aims, and SL3D can be a good research baseline for unsupervised 3D recognition.
Moreover, SL3D generates good clusters of pseudo labels, and its pretrained weights can improve supervised learning via transfer learning. We hope this work can inspire others to do further research in this area.

\bibliographystyle{splncs}
\bibliography{egbib}

\newpage

\large 
\begin{center}
    \textbf{Appendix for SL3D: Self-supervised-Self-labeled 3D Recognition} \
\end{center} 
\appendix

\section{Model Illustrations}
\label{sec:illustration}
\begin{figure*}[!ht]
\includegraphics[width=1.0\textwidth]{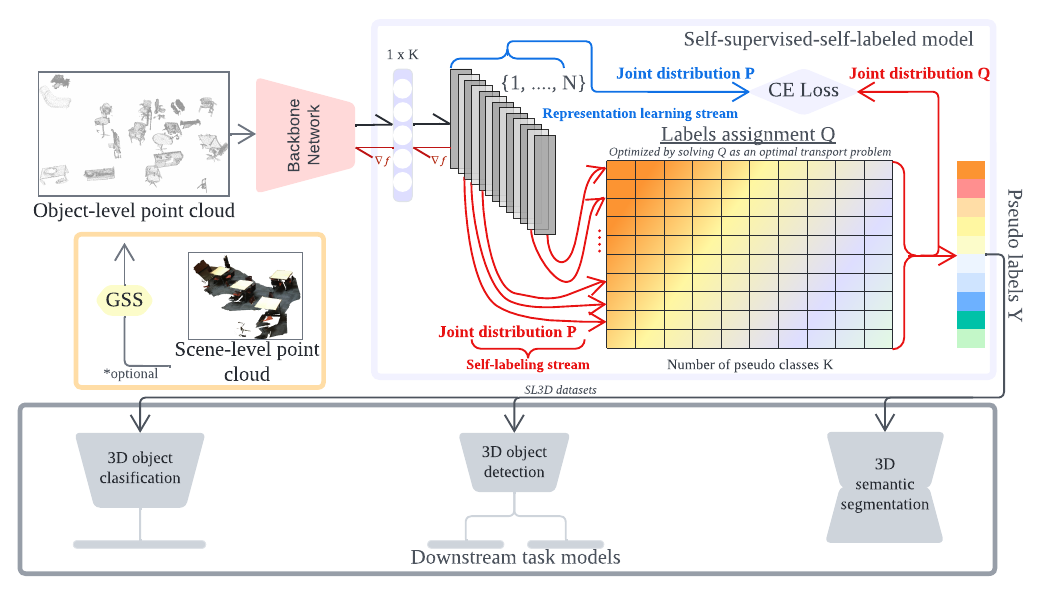}
\caption{Approach overview: Given object-level point cloud, the model simultaneously clusters and learns feature representation by iteratively cluster features to generate pseudo label and use the labels to guide the feature learning. These pseudo labels can then be used to train several downstream tasks such as 3D object classification, 3D object detection, and 3D semantic segmentation. *For scene-level point cloud data, a geometric selective search algorithm \cite{ren2021wypr} is employed to transform scene data into set of object-level point clouds}\label{fig:1} 
\end{figure*}

\section{Related Work}

\label{sec:related}

In the following, we  discuss and highlight existing works for both 3D self-supervised learning and 3D object recognition.

\subsection{Self-supervised Learning}
A deep learning approach that does not require any supervision signal during learning \cite{caron2020unsupervised,DBLP:journals/corr/abs-1911-05722,caron2018deep,DBLP:journals/corr/DoerschGE15,DBLP:journals/corr/abs-1805-00385,DBLP:journals/corr/ZhangIE16,larsson2017colorization,jenni2018selfsupervised,noroozi2017representation,chen2020mocov2,asano2020self,gidaris2018unsupervised,ji2019invariant}. 
Instead, it learns visual representations from a given unlabeled dataset. There are several approaches related to 3D self-supervised learning. The contrastive learning approach learns to compare two augmentations of one input sample, and minimizes the embedding distance for features from the same sample and maximizes the embedding distance for those belonging to different samples. Xie \textit{et al.} \cite{https://doi.org/10.48550/arxiv.2007.10985} propose the first transfer learning method for 3D scene understanding by applying contrastive learning on point clouds. Li \textit{et al.} \cite{https://doi.org/10.48550/arxiv.2110.08188} improve semi-supervised segmentation model by utilizing unlabeled data to enhance feature learning. Recently, for self-supervised representation learning from point clouds, different pretext tasks have been exploited to leverage the properties of 3D point cloud. Sauder and Sievers \cite{https://doi.org/10.48550/arxiv.1901.08396} design a model that learns to reconstruct the original point cloud data from randomly rearranged input points for representation learning. Poursaeed \textit{et al. }\cite{poursaeed2020selfsupervised} learn to predict the 3D rotation of objects. Thabet et al. \cite{thabet2019mortonnet} design a recurrent neural network model to predict the next point from a given point sequence generated by a fast space-filling z-order curve. Recently, multi-modality information is also explored in  \cite{https://doi.org/10.48550/arxiv.2203.00680} for 3D representation learning by maximizing the agreement in the latent space between point cloud and its 2D correspondence.

The above efforts aim to learn visual representations from  given unlabeled data, which can later be used as a pretrained model for further fine-tuning on downstream tasks to alleviate the data requirement. However, in order to solve a 3D recognition task, annotation data is still required in the fine-tuning stage. In contrast with most 3D self-supervised learning models, our framework can

self-label data with clustering and learn visual representations  from pseudo labeled data. 
In this way, not only can the pre-trained model weights be used for further fine-tuning downstream tasks, but we can also utilize the obtained pseudo labels to solve several 3D recognition tasks.

\subsection{3D Object Recognition} 

3D object recognition is one fundamental problem in computer vision that aims to process given 3D visual data to generate high-level understandings. \textit{3D object classification}, \textit{3D object detection}, and \textit{3D semantic segmentation} are three representative tasks.

{3D object classification} attempts to recognize the class label for a particular object-level point cloud. Early attempts to solve this problem by recognizing multi-view images  \cite{su2015multi,yu2018multi,yavartanoo2018spnet} or applying a 3D convolutional network on voxelized 3D data \cite{maturana2015voxnet,wu20153d}. To avoid generating multi-view images or voxels in the 3D space, PointNet \cite{qi2017pointnet} is proposed by Qi \textit{et al.} that directly learns from unordered object-level point cloud data. PointNet++ \cite{qi2017pointnetplusplus} further enhances the previous model by introducing hierarchical feature learning which improves its robustness towards point density variations. The current state of the art 3D object classification, PointMLP \cite{ma2022rethinking}, uses a pure residual MLP network and a lightweight geometric affine module. Graph-based learning \cite{wang2019dynamic,li2019deepgcns}, dynamic neural network~\cite{xu2021paconv}, and attention architecture \cite{zhao2021point,guo2021pct} are also 
investigated for point cloud classification.

{3D object detection} focuses on classifying and locating objects inside a scene-level point cloud by estimating each object's oriented 3D bounding boxes. Upon a PointNet++ backbone for feature extraction, VoteNet \cite{qi2019deep} designs a voting mechanism to improve bounding box proposal qualities. VoteNet is a representative work in 3D object detection. Recently, unlike \cite{qi2019deep}, RBGNet \cite{wang2022rbgnet} utilizes a ray-based feature grouping module to improve the 3D bounding box proposal qualities.  Weakly-supervised learning on 3D object detection tasks has also been investigated such as \cite{ren2021wypr,xu2022back}, which leverage weak annotations as supervision.

{3D semantic segmentation} focuses on point-level semantic prediction. 3D semantic segmentation methods can be further categorized into point-based approaches \cite{qi2017pointnet,qi2017pointnetplusplus,zhao2021point,xu2021paconv}, which directly extract features from raw  input point cloud data and volumetric-based approaches \cite{choy20194d,graham20183d,hu2021bidirectional,hu2020jsenet} , which employ a 3D structured CNN to extract voxelized discrete grid input data. 
Data augmentation strategies such as Mix3D \cite{nekrasov2021mix3d} are also proposed to improve semantic segmentation performance. Recently, weakly supervised learning is also investigated in point cloud segmentation, which learns segmentation from weak labels such as scene category labels \cite{ren2021wypr} and sparse point annotations \cite{liu2021one}.

These proposed methods all focus on either learning a model from annotated training data or using a self-supervised learning method to improve model feature learning. In contrast, we develop an unsupervised learning method for 3D recognition tasks. Moreover, in our work, we adopt PointNet++ \cite{qi2017pointnetplusplus} method for both point cloud classification and segmentation and VoteNet \cite{qi2019deep} method for point cloud object detection. 

\section{Implementation Details}
\label{sec:implementation}
This subsection explains the details of each component in SL3D model, i.e., GSS, Self-supervised-Self-label model, pseudo class annotation, and the downstream task models.

\paragraph{Geometric selective search (GSS). }GSS object proposal algorithm is divided into two steps: \textit{Step 1: } detect basic geometry shapes of given scene-level point clouds and \textit{Step 2: } group similar shapes over a region hierarchically based on a similarity score to generate 3D proposals. For detecting geometry shapes, we use the efficient and reliable Computational Geometry Algorithms Library (CGAL) \cite{fabri2009cgal}. Specifically, we use the region-growing based algorithm in the shape detection library. We set the parameters of the region-growing based algorithm with a search space of $12$ nearest neighbors, the maximum accepted angle between point's normal and plane's normal as $20^{\circ}$, and the minimum region size to $50$. For the second step, i.e., the hierarchical agglomerative clustering algorithm, we select size and volume similarity score to generate the 3D bounding box proposals following settings in \cite{ren2021wypr}. Moreover, we apply random jitters for every $P^S_{i}$  before computing convex hull to improve the proposals recall rate and set the maximum number of proposals to be $1000$. Finally, we use the NMS algorithm with an IoU threshold of $0.75$ to remove redundant bounding boxes generated by GSS, with all bounding boxes post-processed by the NMS algorithm having a number of point sets smaller than $15k$ points. 

\paragraph{Self-supervised-Self-labeled model. } SL3D uses two popular backbone networks, i.e., PointNet++ \cite{qi2017pointnetplusplus} and Point Transformer \cite{zhao2021point}, where the input data with $2048$ points of object-level point cloud $P^O_{i}$ is sampled to $1024$ points, and the network output a $K\times 1$ feature vector for each object. We train our model for $600$ epochs with batch size $32$, a learning rate with an initial value of $0.001$, and decayed $10\times$ for every $200$ epochs. 
For optimization, a stochastic gradient descent optimizer is used with a momentum of $0.9$ and a weight decay of $0.0001$. For the Sinkhorn-Knopp (SK) algorithm, we set $\lambda$  to be 25 and the number of optimization to be $100$. During our model training, we set several pseudo label's numbers $\{18, 40, 50, 100, 200, 400\}$. Our model is computed using $4$ NVIDIA GeForce RTX 3090 GPUs, with each unit equipped with $24$ Gigabyte memory size. 

\begin{wrapfigure}{r}{5.5cm}
\label{wrap-fig:1}
\includegraphics[width=5.5cm]{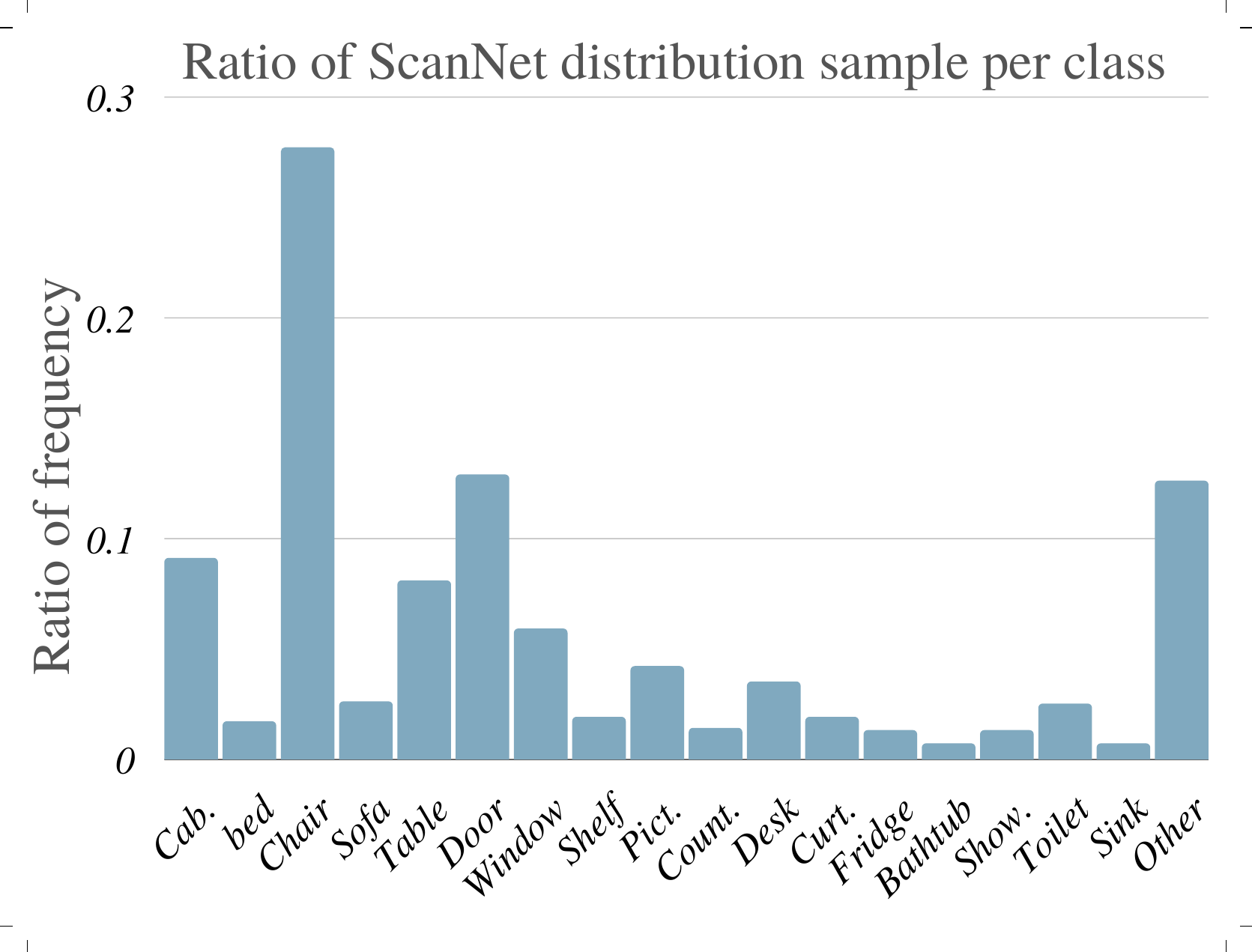}
\caption{Overview of the class distribution on ScanNet dataset (combined training and validation sets) *without floor and wall classes.}
\end{wrapfigure}

\begin{figure*}[ht]
\label{fig:3}
\includegraphics[width=0.94\textwidth]{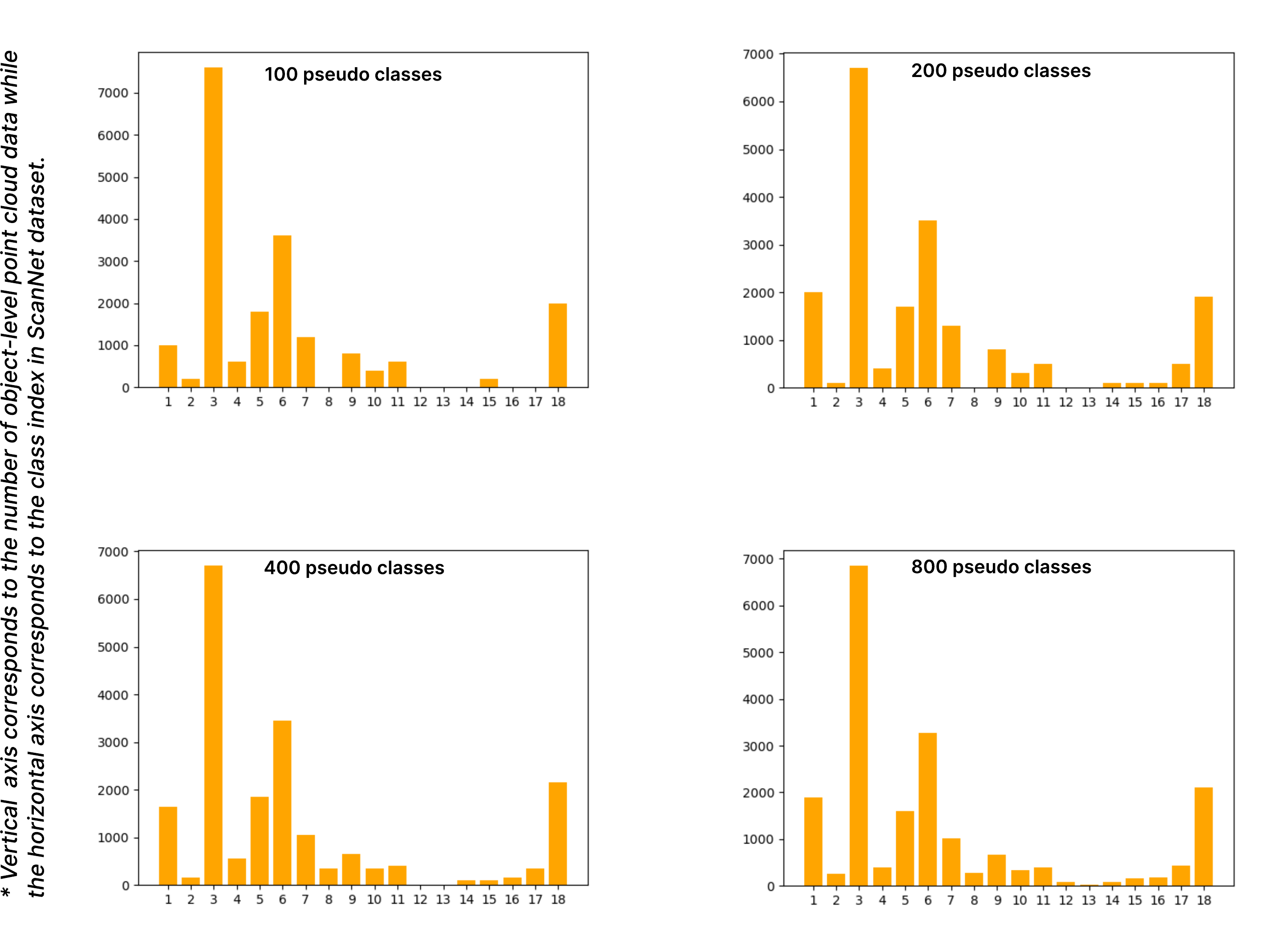}
\caption{Additional visualization for ScanNet class distribution comparison between different number of pseudo classes.}
\end{figure*}

\begin{figure*}[ht]
\label{fig:4}
\includegraphics[width=0.94\textwidth]{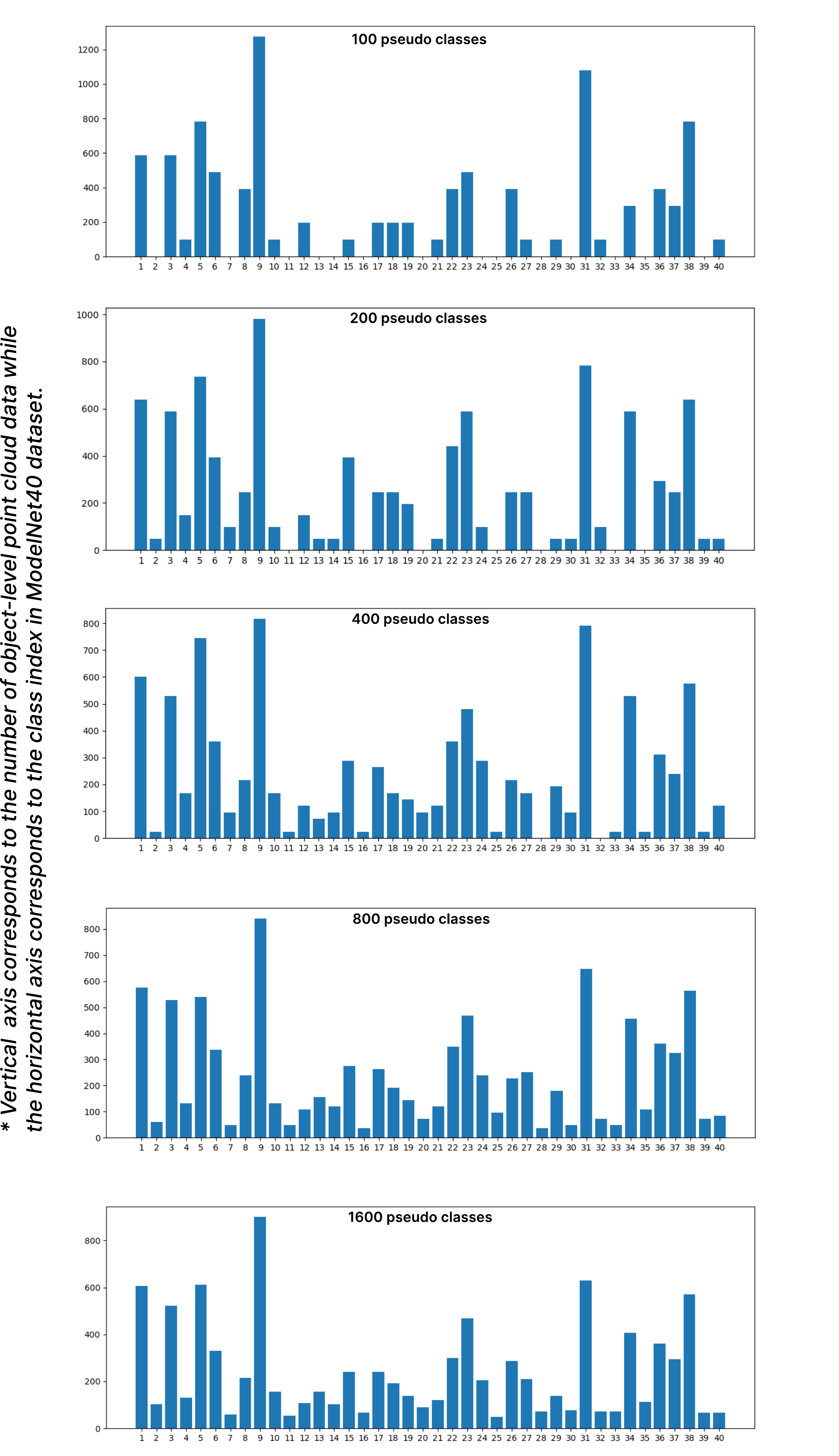}
\caption{Additional visualization for ModelNet40 class distribution comparison between different number of pseudo classes.}
\end{figure*}

\paragraph{Pseudo class annotations. }To evaluate our model on the  standard object recognition benchmark, we need to align the object groups discovered by our model with the number of ground-truth classes. We have to align each of our datasets (depending on the task) by manually group similar pseudo classes into one class label. 
As shown in Table~\ref{table:1}, SL3D requires a large number of pseudo classes to discover all ScanNet semantic classes. There are three reasons behind this. First, it is due to the class imbalance problem in ScanNet dataset, as shown in Fig. \hyperref[wrap-fig:1]{2}. In total, ScanNet dataset has $20,096$ object-level point clouds from training ($\sim78\%$) and validation ($\sim22\%$) sets where $27.7\%$ of the objects belong to ``chair''. Second, there is a possibility that the object-level point cloud generated by the GSS algorithm might contain noise, e.g., backgrounds. Therefore, these pseudo classes do not have any corresponding category in the ground truth data. 
Lastly, SL3D has an equipartition constraint that forces each pseudo class to have an equal number of data. To better understand the influence of pseudo class number towards the actual class distributions, We also shown the comparison of class distribution after grouping process generated from different number of pseudo classes as shown in Fig. \hyperref[fig:3]{3} and Fig. \hyperref[fig:4]{4} for both ScanNet and ModelNet datasets respectively.

\begin{table}[ht]
\caption{Classes alignment between pseudo labels and ScanNet semantic classes (18 object classes + wall + floor). $\dagger$ denotes SL3D with PointNet++ backbone network while $\ddagger$ denotes SL3D with Point Transformer backbone network.}
\centering
\begin{tabular}{c | c c c c c c c c c c}

\toprule

 {No. pseudo classes} & {$18^\dagger$} & {$18^\ddagger$} & {$50^\dagger$} & {$50^\ddagger$} & {$100^\dagger$} & {$100^\ddagger$} & {$200^\dagger$} & {$200^\ddagger$} & {$400^\dagger$} & {$400^\ddagger$} \\  
 {No. obtained classes} & {$7$} & {$9$} & {$9$} & {$11$} & {$13$} & {$14$} & {$16$} & {$17$} & {$18$} & {$20$} \\ 
\bottomrule
\end{tabular}
\label{table:1}
\end{table}

\vspace{-5mm}

\paragraph{Downstream task models. } The downstream task model for both 3D object detection and 3D semantic segmentation can can be any model architecture taking point cloud as inputs. Here, we adopt the classic network for specific tasks.

For the \textit{3D object classification task}, we train PointNet++ \cite{qi2017pointnetplusplus} classification model with SL3D pseudo labeled data.

For the \textit{3D object detection task}, the VoteNet \cite{qi2019deep} model is adopted. which takes as inputs raw scene-level point cloud data and is supervised by pseudo labels generated by SL3D. To tailor the detection model for evaluation, an alignment process is conducted to map pseudo labeled data to the corresponding ground-truth categories.
We train the model for $180$ epochs with a batch size of $32$, a learning rate with an initial value of $0.001$, and is decayed $10\times$ at epochs $\{80, 120, 160\}$. 
To optimize the model, an Adam optimization algortihm is used without weight decay.

For the \textit{3D semantic segmentation task}, we choose PointNet++ segmentation model for semantic scene labeling with Multi Scale Grouping (MSG) settings \cite{qi2017pointnetplusplus}. The model takes as inputs raw scene-level point cloud data and is supervised by pseudo labels from \textit{SL3D}. We train PointNet++ segmentation model for $500$ epochs with $32$ batch size, a learning rate with an initial value of $0.001$, and is decayed $1.5\times$ for every $100$ epochs. An Adam optimizer is used with no weigt decay to optimize the model training.

\section{Ablation Studies}
\label{sec:ablation}

In order to better understand our model, we perform ablation studies on individual components of our model. As GSS \cite{ren2021wypr} is not our contribution, the ablation study of GSS is not included in this ablation study.
In the following, we first study the clustering quality of SL3D. Then investigate whether the pretrained SL3D model can learn meaningful representations using supervised finetuning and k-NN evaluation \cite{caron2018deep}. All ablation studies are performed on the ScanNet dataset.
\subsection{Clustering Evaluation} 

We use ScanNet ground-truth instance object-level point cloud data (training and validation sets) and generate pseudo labels from it using our self-supervised-self-labeled model. To analyze the clustering quality, we use mean purity evaluation measurement  For each group, the purity ratio is calculated as the number of objects belonging to the majority class over the total number of data samples.  The mean purity ratio is the average of purity ratio over all categories. This is not a perfect metric as it might lead to a trivial solution as purity value of $100\%$ can be achieved by setting the number of pseudo classes to be equals with the total number of data. Considering that the total number of evaluated objects is 20,096, which is much larger than pseudo class number, the mean purity ratio can be used as a proxy metric. We test the model using both PointNet++ and Point Transformer backbone networks with $\{18,50,100,200,400\}$ number of pseudo-classes as shown in Table \ref{table:5}.

\begin{table}[ht]
\caption{Clustering quality (\%) on SL3D dataset.}
\centering
\begin{center}
\begin{tabular}{c c | p{1cm} p{1cm} p{1cm} p{1cm} p{1cm}} \toprule
    \multicolumn{1}{c}{\multirow{2}{*}{No.}}&\multicolumn{1}{c |}{\multirow{2}{*}{Backbone}}&\multicolumn{ 5}{c}{\textbf{No. SL3D pseudo classes}}\\
    \cline{3-7}
    \multicolumn{2}{c |}{} & {18} & {50} & {100} & {200} & {400}\\ \midrule
    {1}  & {PointNet++ \cite{qi2017pointnetplusplus}} & {40.6} & {43.5}  & {45.6} & {47.3} & {53.9}\\
    {2}  & {Point Transformer \cite{zhao2021point}}  & {42.5} & {47.1}  & {49.9} & {55.2} & {61.1}\\ \bottomrule
\end{tabular}
\end{center}
\label{table:5}
\end{table}

Table \ref{table:5} shows that our model performs reasonably well  under different settings, manifesting the quality of clustering results. The Point Transformer backbone delivers the best result of $61.1\%$ of mean purity ratio  with $400$ pseudo classes. And the PointNet++ backbone yields a purity ratio of  $53.9\%$ with $400$ pseudo classes 

\subsection{Analysis on Pretrained Weights through Finetuning}
The purpose of conducting the  finetuning evaluation is to assess whether our pretrained model through SL3D can learn useful representations. 
First, we trained our self-supervised model with the object-level point cloud data generated by GSS. 
Second, the pre-trained model weights are used as weight initialization and finetuned on ground-truth object-level point cloud. We use ScanNet object-level point cloud dataset for model training and testing, where $15,733$ objects are allocated for training set and $4,363$ objects for testing set.

As shown in Table \ref{Table:6}, we compare two types of model configurations. The first model is the one that uses random initialization, while the second model uses our SL3D pretrained weight as its weight initialization. Moreover, we also show the performance difference when a different type of backbone network is used. We observe that the second model outperforms the first model under all settings (e.g., type of backbone network, number of pseudo classes). This observation shows that the quality of our SL3D’s learned representation features helps model to generalize and perform better.

\begin{table}[ht]
\caption{\label{Table:6} Finetuned and k-NN evaluations on SL3D pretrained weights}
\centering
\begin{tabular}{c c c c c | c} \toprule
    \multicolumn{1}{c}{\multirow{3}{*}{No.}}&\multicolumn{1}{c}{Weight init.}&\multicolumn{1}{c}{No. of pseudo }&\multicolumn{1}{c}{Finetuned eval.}&\multicolumn{2}{c}{k-NN eval.}\\
    \cmidrule(lr){4-4} \cmidrule(lr){5-6}
    \multicolumn{1}{c}{} & \multicolumn{1}{c}{SL3D pretrained} &{classes} & {Acc (\%)} & \multicolumn{1}{c}{20-NN} & \multicolumn{1}{c}{100-NN}\\ 
    \multicolumn{1}{c}{} & \multicolumn{1}{c}{$(\cmark / \xmark)$} & \multicolumn{2}{c}{} & \multicolumn{1}{c}{Top1:Top5} & \multicolumn{1}{c}{Top1:Top5}\\ \midrule
    \multicolumn{6}{S}{\textbf{Backbone: PointNet++ \cite{qi2017pointnetplusplus}}} \\ \midrule
    {1}  & {\xmark} & {-} & {82.14} & {44.76 : 85.40} & {42.29 : 88.12}\\
    {2}  & {\cmark} & {18} & {83.76}  & {48.20 : 86.53}  & {45.73 : 88.85}\\
    {3}  & {\cmark} & {50} & {82.92} & {48.40 : 86.81}  & {48.20 : 89.61}\\
    {4}  & {\cmark} & {100} & {83.46} & {48.88 : 86.95}  & {48:49 : 89.72}\\
    {5}  & {\cmark} & {200} & {83.95} & {50.24 : 87.64}  & {49.76 : 90.06}\\
    {6}  & {\cmark} & {400} & {83.44} & {52.66 : 88:08}  & {51.48 : 90:31}\\ \midrule
    \multicolumn{6}{S}{\textbf{Backbone: Point Transformer \cite{zhao2021point}}} \\ \midrule
    {7}  & {\xmark} & {-} & {79.64} & {46.27 : 85.83}  & {45.79 : 88.15}\\ 
    {8}  & {\cmark}  & {18} & {82.25} & {59.71 : 93.10} & {55.74 : 93.87}\\
    {9}  & {\cmark}  & {50} & {81.54} & {60.98 : 93.17}  & {57.01 : 93.64}\\
    {10}  & {\cmark}  & {100} & {82.35} & {66.67 : 93.71}  & {63.14 : 94.57}\\
    {11}  & {\cmark}  & {200} & {82.60} & {69.05 : 95.08}  & {66.67 : 95.49}\\
    {12}  & {\cmark}  & {400} & {80.51} & {71.34 : 95.39}  & {69.31 : 96.22}\\ \bottomrule

\end{tabular}
\end{table}

\subsection{Analysis on k-NN Evaluation Results}

We also experiment on our pretrained model using k-Nearest Neighbour to evaluate the discriminativeness of our features. We divide our experiment into two configurations. The first configuration uses a randomly initialized weight, while the second  initializes its weight using SL3D pretrained weights. We evaluate the k-NN model (embedding size of 1024) using ScanNet object-level point cloud data, where $15,733$ objects are allocated for training set and $4,363$ objects for testing set.
 
The k-NN evaluation result is shown in Table \ref{Table:6}, and we show that our SL3D pretrained weights to a great extent help the k-NN model to achieve much higher accuracy and outperform the k-NN model with randomly initialized weight in all settings. We further analyze our pretrained weight by comparing it with a different types of backbone networks. For a randomly initialized weight k-NN model, the 20-NN and 100-NN settings with different backbone networks have accuracy differences (Top 1 : Top 5) of $(1.51:0.43)$ and $(3.5:0.03)$, respectively, where the model with Point Transformer performs slightly better. With our pretrained weights, the accuracy differences become $(15.87 \pm 3.2:6.88 \pm 0.42)$ and $(13.64 \pm 3.63:5.04 \pm 0.63)$, respectively. This significant increase shows that SL3D pretrained weights with Point Transformer as its feature extractor effectively helps model learn discriminative features.

\section{Limitations and Future Directions}
\label{sec:limit}
Our research found that SL3D will encounter an issue when the input dataset is imbalanced. This problem occurred as SL3D's self-labeled mechanism has a constraint where the output label classes must have an equally partitioned number of input datasets among the pseudo-classes. 

\end{document}